\documentclass[runningheads]{llncs}
\usepackage{graphicx}
\usepackage[ruled,vlined]{algorithm2e}
\usepackage[cmex10]{amsmath}
\usepackage{amssymb}
\usepackage{multirow}
\usepackage[misc,geometry]{ifsym}
\begin{document}
\title{Unknown-box Approximation to Improve Optical Character Recognition Performance}
\titlerunning{Unknown-box Approx. to Improve OCR Performance}
%
\author{Ayantha Randika\inst{1} (\Letter)\and
Nilanjan Ray\inst{1} \and
Xiao Xiao\inst{2} \and
Allegra Latimer\inst{2}}
\authorrunning{A. Randika et al.}

\institute{
University of Alberta, 116 St and 85 Ave, Edmonton, AB, Canada
\email{\{ponnampe,nray1\}@ualberta.ca}\and
Intuit Inc., 2700 Coast Ave, Mountain View, CA USA\\
\email{\{Xiao\_Xiao,Allegra\_Latimer\}@intuit.com}
}
\maketitle              
%
\begin{abstract}
Optical character recognition (OCR) is a widely used pattern recognition application in numerous domains. There are several feature-rich, general-purpose OCR solutions available for consumers, which can provide moderate to excellent accuracy levels. However, accuracy can diminish with difficult and uncommon document domains. Preprocessing of document images can be used to minimize the effect of domain shift. In this paper, a novel approach is presented for creating a customized preprocessor for a given OCR engine. Unlike the previous OCR agnostic preprocessing techniques, the proposed approach approximates the gradient of a particular OCR engine to train a preprocessor module. Experiments with two datasets and two OCR engines show that the presented preprocessor is able to improve the accuracy of the OCR up to 46\% from the baseline by applying pixel-level manipulations to the document image. The implementation of the proposed method and the enhanced public datasets are available for download\footnote{https://github.com/paarandika/Gradient-Approx-to-improve-OCR}.

\end{abstract}

\section{Introduction}
Optical Character Recognition (OCR) is the process of extracting printed or handwritten text from an image into a computer understandable form. Recent OCR engines based on deep learning have achieved significant improvements in both accuracy, and efficiency \cite{Chen2019}. Thanks to these improvements, several commercial OCR solutions have been developed to automate document handling tasks in various fields, including cloud-based OCR services from big names in cloud technologies. This new generation of OCR solutions come with all the benefits of the Software as a Service (SaaS) delivery model where the consumer does not have to think about the hardware and maintenance.

One drawback of commercial OCR solutions is that they are often general-purpose by design, while different domains may have different document types with unique aberrations and degradations, which can hinder OCR performance. While it is theoretically possible to either train an open-source OCR or to build a new OCR engine from scratch to accommodate the domain shift, these solutions are not realistic in practice for OCR users in the industry, given the amount of resources required for training and the potential degradation of efficiency. Therefore, the more viable approach is to enhance a commercial OCR solution with pre- or post-processing to improve accuracy.

The output of OCR is a text string and can be post-processed using natural language-based techniques \cite{Thompson2016} or by leveraging outputs from multiple OCR engines \cite{Reul2018}. This is beyond the scope of this work. Instead, we focus on preprocessing to enhance input quality prior to the OCR, which usually occurs in the image domain, though earlier hardware-based preprocessors have worked in the signal domain \cite{White1983,Hicks1979}. Image preprocessing for OCR includes image binarization, background elimination, noise removal, illumination correction and correcting geometric deformations, all of which aim to produce a simpler and more uniform image that reduces the burden on the OCR engine \cite{Sauvola2000}.

In this work, we propose a preprocessing solution which performs pixel-level manipulations that can be tweaked to accommodate any OCR engine, including commercial `unknown-box' OCRs, in which the components are not open-sourced or known to us. To demonstrate the ability of our preprocessor, we use Point of Sales (POS) receipts, which often have poor printing and ink quality, and therefore renders a challenging task for OCR software \cite{Huang2019}. Additionally, we include the VGG synthetic word dataset \cite{Jaderberg14c} to establish the generality of our solution. Section 2 discusses existing preprocessing techniques used to improve OCR performance. Section 3 presents our approach to the problem, and Section 4 contains the implementation and evaluation of our method. Section 5 discusses the experiment results, and Section 6 concludes the article with future work.

\section{Background}
Image binarization, which converts a grayscale image into black and white, is one of the most widely used preprocessing techniques for OCR since the 1980s. Otsu \cite{otsu1979} is a popular binarization method, which uses the grayscale histogram to find a threshold value. Object Attribute Thresholding \cite{Liu1993} uses grayscale and run-length histograms for unconstrained document images with complex backgrounds. O'Gorman proposed a global thresholding method using local connectivity details \cite{Ogorman1994}. Similarly, a five-step document binarization method \cite{Chang1995} focuses on connectivity issues in characters. Sauvola and Pietikäinen proposed an adaptive binarization method \cite{Sauvola2000} which classifies the image into subcomponents and assigns different threshold values to pixels based on the component types. The double thresholding method developed by Chen et al. uses edge information to generate a binary image \cite{Chen2008}. Some more recent binarization techniques incorporate deep learning. For example, DeepOtsu \cite{He2019} applies Convolutional Neural Networks (CNN) iteratively, while Vo et al. \cite{Vo2018} adopt a CNN-based hierarchical architecture.

Skeletonization is another popular preprocessing technique \cite{lam1995}, which aims to reduce the dimensions of an object \cite{saha2017}. In the context of characters, skeletonization reduces the stroke thickness to 1-D curves. Similar to binarization, there are several existing methods for Skeletonization \cite{saha2017}, some of which are specifically designed for OCR \cite{lam1995}. Other preprocessing approaches exist in addition to binarization and skeletonization. Bieniecki et al. proposed methods to correct geometrical deformations in document images \cite{Bieniecki2007}. An independent component analysis-based method has been developed for handheld images of inscriptions \cite{Garain2008}. Harraj and Raissouni combined different image enhancement techniques with Otsu binarization \cite{Harraj2015}. Deep learning-based Super-Resolution (SR) is employed in \cite{lat2018} and \cite{Peng2020}. Sporici et al. presented a CNN-based preprocessing method where convolution kernels are generated using Reinforcement Learning (RL) \cite{Sporici2020}.

Different preprocessing methods address different shortcomings in incoming images for OCR engines. Binarization and skeletonization focus on creating simpler images, while techniques such as super-resolution attempt to add more details. Geometrical correction methods create uniform characters in shape and orientation. The shared goal of these different methods is to produce output images which the OCR engine is more comfortable or familiar with. However, OCR engines have different underlying architectures and mechanisms, which may lead to different expectations for incoming images or tolerance for various defects. Therefore, we hypothesize that a preprocessor optimized for a given OCR engine would be able to produce a better approximation of the optimal images expected by this specific engine, leading to higher OCR accuracy compared to traditional generic preprocessing methods. 

\section{Proposed Method}

The straightforward way to optimize the parameters of a preprocessor, which manipulates images to approximate the optimal input distribution for the OCR, is to use the optimal input distribution as ground truth. However, such intermediate training data are rarely available. Another way of optimizing the parameters of the preprocessor is by calculating the gradient of the OCR error and propagating it to the preprocessor using the backpropagation algorithm and updating the preprocessor parameters to minimize the error. However, the problem with this approach is that the internal mechanisms of the proprietary OCR engines cannot be accessed. Further, OCR engines may contain non-differentiable components for which gradient computation or estimation may not be possible. Therefore, the OCR engine needs to be treated as an unknown-box in the training pipeline. 

Since there is no direct way of calculating the gradient of an unknown-box, to utilize the backpropagation algorithm to train the preprocessor component, the gradient of the error produced by the input needs to be estimated. A popular method for estimating the gradient is using the Score-Function Estimator (SFE) (\ref{score_fn}) \cite{Mohamed2019}:
\begin{equation}
\label{score_fn}
\nabla_\theta\mathbb{E}_{x \sim p_\theta}[f(x)] = \mathbb{E}_{x \sim p_\theta}[f(x)\nabla_\theta \log{p_\theta(x)}],
\end{equation}
\noindent where $f(x)$ is the unknown-box function and $p_\theta$ is the input distribution parameterized by $\theta$. In RL, (\ref{score_fn}) was developed into the REINFORCE algorithm \cite{Williams1992}. Even though this estimator is unbiased, it can have a high variance, especially in higher dimensions \cite{Mohamed2019}. By re-parameterizing $x$ as $x=\theta+\sigma\epsilon$ where $\epsilon \sim \mathcal{N}(0, \mathcal{I})$, the score function can be written in the following format \cite{salimans2017}:
\begin{equation}
\label{score_fn_2}
\nabla_\theta\mathbb{E}_{\epsilon \sim \mathcal{N}(0, \mathcal{I})}[f(\theta+\sigma\epsilon)] = \frac{1}{\sigma}\mathbb{E}_{\epsilon \sim \mathcal{N}(0, \mathcal{I})}[f(\theta+\sigma\epsilon)\epsilon].
\end{equation}

\begin{algorithm}[t]
    \SetKwInOut{Input}{input}
    \Input{$\sigma, S, \{Training\; Images, Ground\ Truths\}$}
    \For{$I,p_{gt} \in \{Training\; Images,\; Ground\; Truths\}\;$}{
        $g = Preprocessor(I)$\;
        intialize $s$ to 0\;
        \While{$s < S$}{
            sample $\epsilon_s \sim \mathcal{N}(0, \sigma)$\;
            $\mathcal{M}_s = \mathcal{M}(Approximator(g+\epsilon_s),\; OCR(g+\epsilon_s))$\;
            $s = s + 1$\;
        }
        $min_{\phi} \sum_{s}\mathcal{M}_s$\;
        $min_{\psi} \mathcal{Q}(Approximator(g), p_{gt})$\;
    }
    \caption{Gradient approximation by NN}
    \label{algo1}
\end{algorithm}
In the context of medical image analysis, Nguyen and Ray proposed EDPCNN \cite{Nguyen2020} based on the universal function and gradient approximation property of Neural Networks (NN) \cite{Hornik1990UniversalAO}. EDPCNN estimates the gradient of a non-differentiable dynamic programming (DP) algorithm by training a NN as an approximation for the DP algorithm and uses the approximated gradients to train the learning-based component in the processing pipeline. During inference, the approximating NN is removed, and only the non-differentiable original function is used. Jacovi et al. also proposed the same method under the name `Estinet' and demonstrated its performance with tasks such as answering `greater than or less than' questions written in natural language \cite{Jacovi2019}.

In this work, we use the online algorithm proposed by \cite{Nguyen2020} to estimate the gradient of the OCR component. Our approach is detailed in Algorithm \ref{algo1}, which consists of two loops. In the `inner loop', noise is added to `jitter’ the input to the OCR, then the error $\mathcal{M}$ between OCR and the approximator is accumulated as  $\sum_{s}\mathcal{M}_s$. The `outer loop' optimizes approximator parameters ($\phi$) by minimizing the accumulated error $\sum_{s}\mathcal{M}_s$ and freezing the parameters ($\psi$) of the preprocessor. The other minimization in the `outer loop' optimizes the parameters ($\psi$) of the preprocessor model while the approximator parameters ($\phi$) are frozen. For this second minimization, the error $\mathcal{Q}$ is calculated by comparing the approximator output with the ground truth. Note that it is an alternating optimization between the preprocessor and the approximating NN. In its basic components, this algorithm is similar to the DDPG \cite{Lillicrap2016} algorithm in RL.

\section{Implementation and Experiments}
\subsection{Implementation}
\subsubsection{NN-based Approximation Method}
\begin{figure*}[t]
\centering
\includegraphics[width=4.5in]{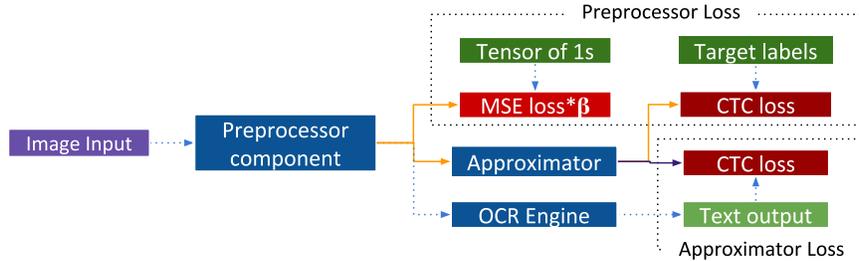}
\caption{Overview of the proposed training pipeline. The yellow and purple arrows represent the computation paths equipped with backpropagation of preprocessor loss and approximator loss, respectively.}
\label{fig_overview}
\end{figure*}

The overview of the proposed training pipeline based on Algorithm \ref{algo1} is depicted in Figure \ref{fig_overview}. The loss function used for optimization of approximator parameters is Connectionist Temporal Classification (CTC) loss \cite{Graves2006}, which appears as $\mathcal{M}$ in Algorithm \ref{algo1}. In addition, mean square error (MSE) loss is calculated by comparing the preprocessor output with a 2-dimensional tensor of ones: $J_{m\times n}$ where $n$ and $m$ are the dimensions of the input image. In this context, it represents a completely white image. Sum of the CTC loss and the MSE loss is used as the loss function (\ref{loss_fn}) to optimize the preprocessor parameters ($\psi$) in Algorithm \ref{algo1}.
\begin{equation}
\label{loss_fn}
\mathcal{Q} = CTC(Approximator(g), p_{gt})+\beta*MSE(g, J_{m\times n}).
\end{equation}
In  (\ref{loss_fn}), $g = Preprocessor(Image)$ and $p_{gt}$ is the associated ground truth text for the input $Image$.  MSE loss component in the loss function nudges the preprocessor to produce a white image. A completely white image implies no output or incorrect output from the approximator, which increases CTC error. We hypothesise that this composite loss function will reduce background clutter while preserving the characters. $\beta$ acts as a hyperparameter to control the effect of MSE loss.

\begin{algorithm}
    \SetKwFunction{FMain}{GetGradients}
    \SetKwProg{Fn}{Function}{:}{}
    \Fn{\FMain{$Image$, $p_{gt}$, $\sigma$, $n$}}{
        $g = Preprocessor(Image)$\;
        sample $\epsilon_1,\dots \epsilon_n \sim \mathcal{N}(0, \mathcal{I})$\;
        $\epsilon_{n+i} = -\epsilon_i$ for $i=1,\dots n$\;
        $\mathcal{L}_i=\mathcal{L}(OCR(g+\sigma\epsilon_i), p_{gt})$ for $i=1,\dots 2n$\;
        $gradient = \frac{1}{2n\sigma}\sum\limits_{i=1}^{2n}\mathcal{L}_i\epsilon_i$\;
        \KwRet $gradient$\;
    }
    \caption{Gradient approximation by SFE}
    \label{algo2}
\end{algorithm}
The architecture of the preprocessor component is based on the U-Net \cite{Ronneberger2015} architecture. We use the U-Net variation \cite{Buda2019} with added batch normalization layers. The number of input and output channels is changed to one since we work with greyscale images. The sigmoid function is used as the final activation function to maintain output values in the range $[0,1]$. Convolutional Recurrent Neural Network (CRNN) \cite{Shi2017} model is used as the approximator to simulate the text recognition of OCR. CRNN model is a simple model, which can avoid gradient vanishing problems when training end-to-end with the preprocessor. An OCR contains different components for text detection, segmentation and recognition. However, CRNN only supports text recognition.

\subsubsection{SFE-based Method}
Since the proposed method is based on gradient estimation, we implement the SFE method (\ref{score_fn_2}) in Algorithm \ref{algo2} that can be viewed as a gradient estimation alternative. Unlike the probability distribution output of CRNN in the NN-based method, OCR outputs a text string. CTC loss is not intended for direct string comparison, and there is no need for a differentiable loss function in SFE; hence, Levenshtein distance \cite{levenshtein1966binary} is used as the loss function, represented by $\mathcal{L}$ in Algorithm \ref{algo2}. This loss is similar to (\ref{cer}), an evaluation metric discussed in section 4.2, except for the multiplication by a constant. $n$ perturbations of $\epsilon$ are sampled from the normal distribution, and the `mirrored sampling’ is used to reduce the variance \cite{salimans2017}. The generated $2n$ samples are sent to the OCR, and the resulting text is evaluated to produce the final loss. The same U-Net model is utilized as the preprocessor, and an adaptation of  (\ref{loss_fn}) is used as the compound loss to optimize it by replacing the CTC loss component with  Levenshtein distance. The gradient of the Levenshtein distance component of the compound loss is calculated with Algorithm \ref{algo2}.

\subsection{Datasets and Evaluation}
Document samples from two different domains are used to evaluate the preprocessor. The dataset referred to as `POS dataset' consists of POS receipt images from three datasets: ICDAR SOIR competition dataset \cite{Huang2019}, Findit fraud detection dataset \cite{Artaud2018} and CORD dataset\cite{Park2019CORDAC}. Since the Findit dataset is not intended as an OCR dataset, ground truth text and word bounding boxes are generated using  Google Cloud Vision API\footnote{https://cloud.google.com/vision/} in combination with manual corrections. Additionally, it contains a fraction of the ICDAR SOIR competition dataset and the entire CORD dataset. Dataset images are patches extracted from POS receipts resized to have a width of $500$ pixels and a maximum of $400$ pixels height. The complete POS dataset contains 3676 train, 424 validation and 417 test images with approximately 90k text areas. The dataset referred to as the `VGG’ dataset is created by randomly selecting a subset of sample images from the VGG synthetic word dataset \cite{Jaderberg14c}. The VGG dataset contains 50k train, 5k validation and 5k test images, each containing a word.

Two free and opensource OCR engines: Tesseract\footnote{https://github.com/tesseract-ocr/tesseract}, a popular well established opensource OCR engine and EasyOCR\footnote{https://github.com/JaidedAI/EasyOCR}, a newer opensource OCR engine are used to train the preprocessor. Throughout the study, both OCR engines are treated as unknown-box components. Two metrics are used to measure the OCR performance variation with preprocessing. Since word-level ground truth values are available for both datasets, word-level accuracy is measured by the percentage of words matched with the ground truth. As a character-level measurement, Levenshtein distance \cite{levenshtein1966binary} based Character Error Rate (CER) is used. CER is defined as:
\begin{equation}
\label{cer}
CER = 100 *(i + s + d) / m,
\end{equation}
where $i$ is the number of insertions, $s$ is the number of substitutions, and $d$ is the number of deletions done to the prediction to get the ground truth text. $m$ is the number of characters in the ground truth. The OCR engine’s performance is measured with original images to establish a baseline, and then the OCR engine is rerun with preprocessed images, and the performance is measured again. 
\subsection{Training Details}

A set of preprocessors are trained for fifty epochs according to the Algorithm \ref{algo1}. Learning rates used for preprocessor and approximator are $5\times 10^{-5}$ and $10^{-4}$ respectively. The approximator is pre-trained with the dataset for fifty epochs before inserting it into the training pipeline to avoid the cold-start problem. Adam optimizer is used in both training paradigms. We maintained $\beta=1$ in (\ref{loss_fn}) and $S=2$ in Algorithm \ref{algo1}. $\sigma$ in Algorithm \ref{algo1} is randomly selected from a uniform distribution containing 0, 0.01, 0.02, 0.03, 0.04 and 0.05. Small standard deviation values for noise are used since the images are represented by tensors in the range $[0.0, 1.0]$. POS dataset images are padded to feed into U-Net. To feed into CRNN, words are cropped using bounding-box values and padded to the size of $128 \times 32$ pixels. Since each VGG sample only contains a single word, the size of $128 \times 32$ pixels is used for both components. On average, it took 19 hours to train a preprocessor with an Nvidia RTX 2080 GPU, while EasyOCR was running on a different GPU.

A different set of preprocessors (referred to as SFE-preprocessor here onwards) are trained with Algorithm \ref{algo2} using Adam optimizer and a learning rate of $5\times 10^{-5}$. To avoid mirrored gradients cancelling each other at the start, the preprocessor is first trained to output the same image as the input. OCR engine output does not vary much with the perturbations of $\epsilon$ if the $\sigma$ is too small. If the $\sigma$ is too large, every perturbation produces empty text. Both of these scenarios lead to $0$ gradients. By trial and error, a constant $\sigma$ of $0.05$ is used. The speed bottleneck of the pipeline is at the OCR. To reduce the training time, $n=5$ is used. Similar to the previous training scenario, we maintained $\beta=1$ (refer to (\ref{loss_fn})), and the same image processing techniques are used for both datasets. On average, it took 79 hours to train an SFE-preprocessor with an Nvidia RTX 2080 GPU while EasyOCR was running on a different GPU.
\begin{table}
\centering
\caption{Accuracy and CER before and after proposed NN approximation-based preprocessing.}
\label{tab:acc}
\begin{tabular}{|l|l|r|r|r|r|r|r|}
\hline
\multicolumn{1}{|c|}{\multirow{3}{*}{\begin{tabular}[c]{@{}c@{}}OCR used for \\ training and \\ testing\end{tabular}}} &
  \multicolumn{1}{c|}{\multirow{3}{*}{Dataset}} &
  \multicolumn{2}{c|}{Without preprocessing} &
  \multicolumn{4}{c|}{With preprocessing} \\ \cline{3-8} 
\multicolumn{1}{|c|}{} &
  \multicolumn{1}{c|}{} &
  \multicolumn{1}{c|}{Accuracy $\uparrow$} &
  \multicolumn{1}{c|}{CER $\downarrow$} &
  \multicolumn{1}{c|}{Accuracy $\uparrow$} &
  \multicolumn{1}{c|}{CER $\downarrow$} &
  \multicolumn{1}{c|}{\begin{tabular}[c]{@{}c@{}}Accuracy\\ gain\end{tabular}} &
  \multicolumn{1}{c|}{\begin{tabular}[c]{@{}c@{}}CER\\ reduction\end{tabular}} \\ \hline
Tesseract & POS & 54.51\% & 26.33 & 83.36\% & 8.68  & 28.86\% & 17.66 \\ [0.5ex]
Tesseract & VGG & 18.52\% & 64.40 & 64.94\% & 14.70 & 46.42\% & 49.70 \\ [0.5ex]
EasyOCR   & POS & 29.69\% & 44.27 & 67.97\% & 16.46 & 38.27\% & 27.81 \\ [0.5ex]
EasyOCR   & VGG & 44.80\% & 26.90 & 57.48\% & 17.15 & 12.68\% & 9.75  \\ \hline
\end{tabular}
\vspace*{1 cm}
\centering
\caption{Proposed NN approximation-based vs. SFE-based preprocessors.}
\label{tab:acc2}
\begin{tabular}{|l|l|r|r|r|r|r|r|}
\hline
  \multicolumn{1}{|c|}{\multirow{3}{*}{\begin{tabular}[c]{@{}c@{}}OCR used for \\ training and \\ testing\end{tabular}}} &
  \multicolumn{1}{|c|}{\multirow{3}{*}{Dataset}} &
  \multicolumn{2}{c|}{NN-based preprocessing} &
  \multicolumn{2}{c|}{SFE-based preprocessing} \\ \cline{3-6} 
\multicolumn{1}{|c|}{} &
\multicolumn{1}{c|}{} &
  \multicolumn{1}{c|}{\multirow{2}{*}{Accuracy $\uparrow$}} &
  \multicolumn{1}{c|}{\multirow{2}{*}{CER $\downarrow$}} &
  \multicolumn{1}{c|}{\multirow{2}{*}{Accuracy $\uparrow$}} &
  \multicolumn{1}{c|}{\multirow{2}{*}{CER $\downarrow$}}\\
  &&&&&\\\hline
 Tesseract & POS & 83.36\% & 8.68 & 69.17\% & 16.62\\ [0.5ex]
 Tesseract & VGG & 64.94\% & 14.70 & 27.76\% & 52.98 \\ [0.5ex]
 EasyOCR   & POS & 67.97\% & 16.46 & 46.63\% & 28.13\\ [0.5ex]
 EasyOCR   & VGG & 57.48\% & 17.15 & 47.02\% & 24.69\\\hline
\end{tabular}
\vspace*{1 cm}
\centering
\caption{OCR engine accuracy on POS dataset: comparison with other preprocessing methods.}
\label{tab:comp}
\begin{tabular}{|l|r|r|r|r|}
\hline
\multirow{2}{*}{Preprocessing Method} & \multicolumn{2}{c|}{Tesseract}                           & \multicolumn{2}{c|}{EasyOCR}                             \\ \cline{2-5} 
                        & \multicolumn{1}{c|}{Accuracy $\uparrow$} & \multicolumn{1}{c|}{CER $\downarrow$} & \multicolumn{1}{c|}{Accuracy $\uparrow$} & \multicolumn{1}{c|}{CER $\downarrow$} \\ \hline
OCR \emph{(no preprocessing)} & 54.51\% & 26.33 & 29.69\% & 44.27 \\ [0.5ex]
Otsu \cite{otsu1979}     & 50.98\% & 29.84 & 16.96\% & 52.30 \\ [0.5ex]
Sauvola \cite{Sauvola2000}  & 55.39\% & 25.20 & 20.19\% & 48.49 \\ [0.5ex]
Vo \cite{Vo2018}       & 51.72\% & 31.81 & 21.95\% & 50.07 \\ [0.5ex]
robin    & 57.18\% & 28.45 & 27.59\% & 43.55 \\ [0.5ex]
DeepOtsu \cite{He2019} & 62.47\% & 21.88 & 26.33\% & 42.63 \\ [0.5ex]
SR \cite{Peng2020}       & 67.13\% & 15.90 & 37.51\% & 31.11 \\ [0.5ex]
\textbf{Ours}                    & \textbf{83.36\%}              & \textbf{8.68}            & \textbf{67.97\%}              & \textbf{16.46}           \\ \hline
\end{tabular}
\vspace*{1 cm}
\centering
\caption{OCR accuracy when trained and tested with different engines.}
\label{tab:cross}
\begin{tabular}{|l|l|l|r|r|}
\hline
\multicolumn{1}{|c|}{Dataset} &
  \begin{tabular}[c]{@{}l@{}}OCR used\\ for training\end{tabular} &
  \begin{tabular}[c]{@{}l@{}}OCR used\\ for testing\end{tabular} &
  \multicolumn{1}{c|}{Test accuracy $\uparrow$} &
  \multicolumn{1}{c|}{Test CER $\downarrow$} \\ \hline
POS dataset & Tesseract & EasyOCR   & 40.44\% & 31.28 \\ [0.5ex]
VGG dataset & Tesseract & EasyOCR   & 47.14\% & 21.77 \\ [0.5ex]
POS dataset & EasyOCR   & Tesseract & 60.94\% & 21.81 \\ [0.5ex]
VGG dataset & EasyOCR   & Tesseract & 21.64\% & 51.96 \\ \hline
\end{tabular}
\end{table}
\section{Results and Discussion}
Results in Table \ref{tab:acc} and Table \ref{tab:acc2} show that the performances of both OCR engines are improved by preprocessing, and NN approximation-based preprocessors outperform SFE-preprocessors. To some extent, the poor performance of the SFE-preprocessor might be attributed to the small number of perturbations. However, when considering computational time, using large $n$ appears unpragmatic. In both cases, gradient approximation has proved to work, and it appears that the NN approximation-based method handles the image domain better than SFE.

\begin{figure*}[t]
\centering
\includegraphics[width=3.9in]{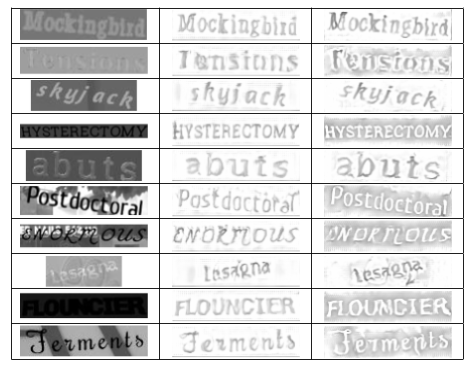}
\caption{Sample inputs and outputs from the VGG test dataset. Column 1: input images, Column 2: image output produced by the model trained with Tesseract, Column 3: image output produced by the model trained with EasyOCR.}
\label{fig_vgg}
\end{figure*}

\begin{figure*}
\centering
\includegraphics[width=4.8in]{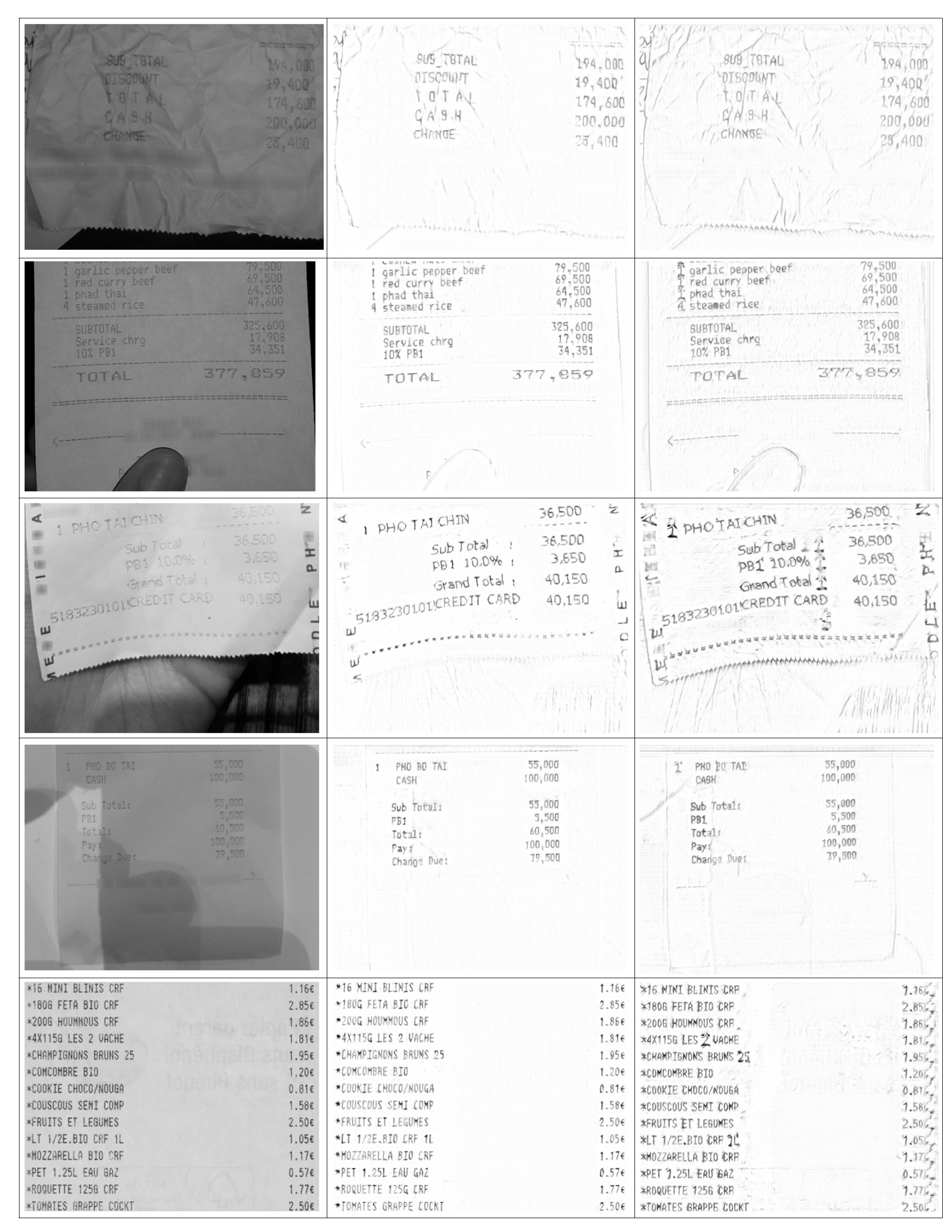}
\caption{Slightly cropped sample inputs and outputs from the POS test dataset. Column one contains the sample input images. Columns two and three contain outputs produced by the proposed NN-based method trained with Tesseract and EasyOCR, respectively.}
\label{fig_pos}
\end{figure*}

Figure \ref{fig_vgg} and Figure \ref{fig_pos} depict sample output images produced by the NN approximation based preprocessors. It can be observed that shadows, complex backgrounds and noise are suppressed to improve the contrast of the text. This provides a clear advantage to the OCR engine. The bleaching of darker text in images can be the effect of MSE loss. Preprocessors have introduced mutations to the characters' shapes, which is more significant in the POS dataset than in the VGG dataset. In the POS dataset, characters have gained more fluid and continuous strokes, especially low-resolution characters printed with visible `dots'. Additionally, the `Tesseract preprocessor' was able to do some level of skew corrections to the text in VGG images. Based on the accuracy improvements and reduction of CER, it can be concluded that these mutations provide extra guidance in character recognition.

\begin{figure*}
\centering
\includegraphics[width=4.3in]{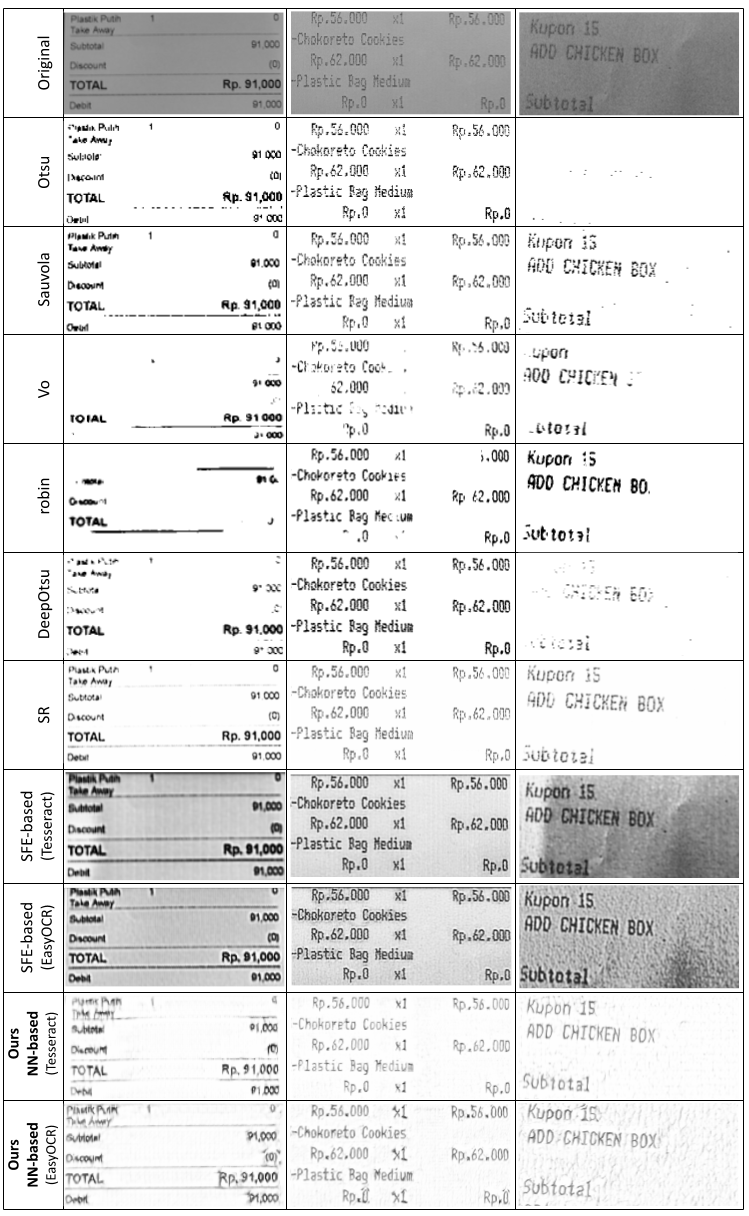}
\caption{Cropped input (POS test dataset) and output images from methods considered in Table \ref{tab:comp} and our models. The images in three columns have average CER reductions of 6.5, 61 and 73.5, respectively, from left to right. CER is based on our NN-based models.}
\label{fig_comp}
\end{figure*}
\begin{figure*}
\centering
\includegraphics[width=4.5in]{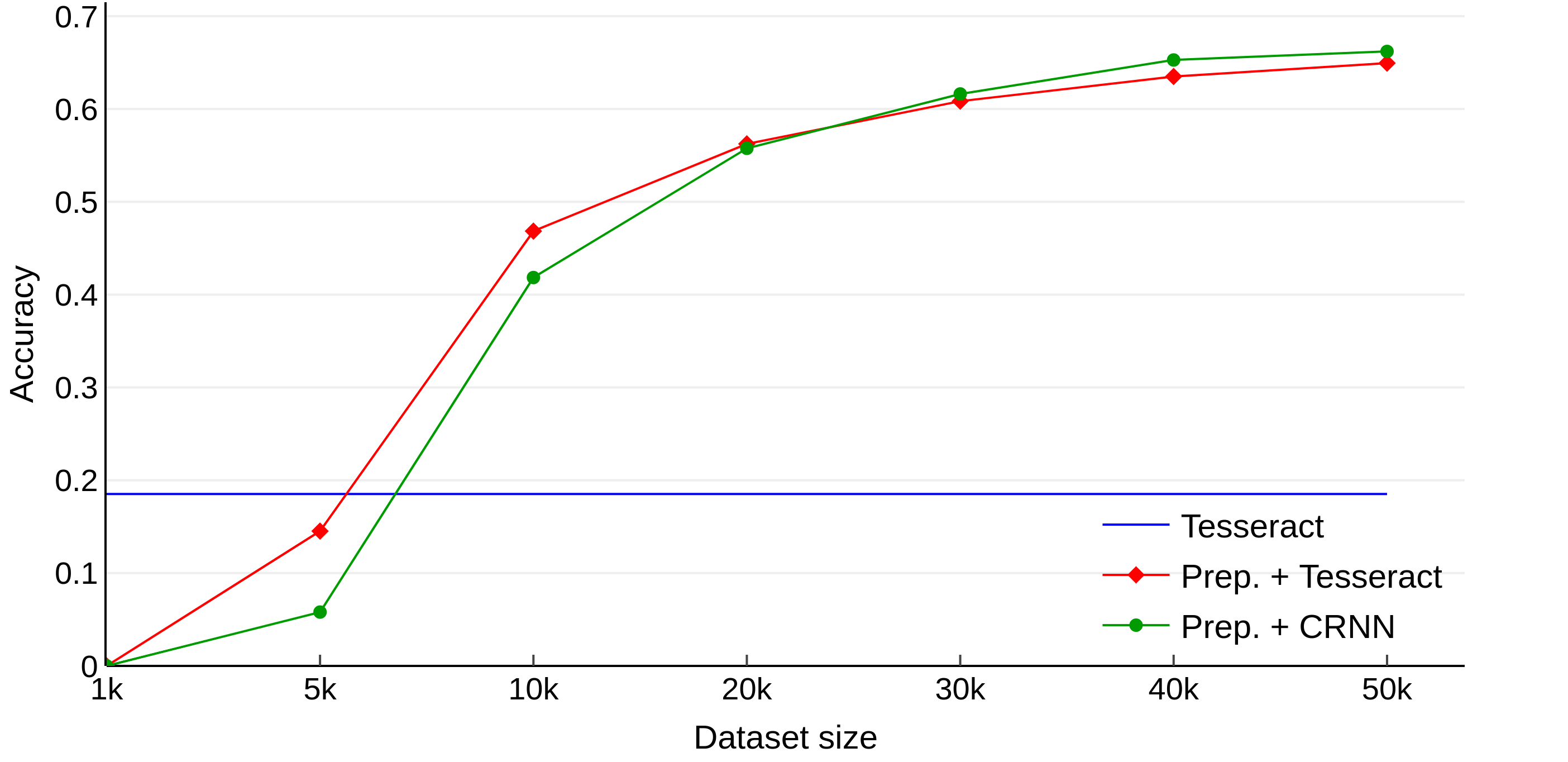}
\caption{The test set accuracy of OCR engine and CRNN model with different training sizes of VGG dataset. Tesseract accuracy without preprocessing is added for reference.}
\label{crnn_comp}
\end{figure*}
Our approach does not require clean document images as labels. The preprocessor is trained directly with the text output omitting the need for intermediate clean data. Due to the lack of clean ground truth images for our dataset, to compare with learning-based methods, we used pre-trained weights. In Table \ref{tab:comp}, five binarization methods and one SR method are compared against our preprocessor on the POS dataset. Vo \cite{Vo2018}, DeepOtsu \cite{He2019} and robin\footnote{https://github.com/masyagin1998/robin} are originally trained with high resolutions images. Therefore, to mitigate the effect of low-resolution, POS dataset images are enlarged by a factor of 2 before binarization and reduced back to the original size before presenting to the OCR engine. Similarly, with SR method \cite{Peng2020}, the images are enlarged by a factor of 2 and presented the same enlarged images to the OCR. According to the results, our method has outperformed all six methods compared. Sauvola, robin, DeepOtsu and SR methods were able to increase the Tesseract accuracy, and the SR method shows the largest improvement. With robin, CER has increased despite the slight accuracy gain. Only the SR method improved EasyOCR accuracy. Furthermore, severe loss of details can be observed in the outputs produced by these methods (Figure \ref{fig_comp}).

To test the effect of individualized preprocessing, preprocessors trained for one OCR engine is tested with a different OCR engine (Table \ref{tab:cross}). After the preprocessing, the OCR engine yields better accuracy than the baseline in the test. However, the accuracy gain is lower than the accuracy obtained by the same OCR engine the preprocessor has trained with. Therefore, it is reasonable to assume that preprocessing has added OCR engine specific artifacts or mutations to the image to improve recognition. This behaviour confirms that different OCR engines expect inputs to be optimized differently, thus individualized preprocessing serves better than generic preprocessing. Additionally, Figure \ref{crnn_comp} shows that the CRNN model has been able to well approximate the recognition capability of the OCR with different dataset sizes. However, given that the CRNN model can only recognize text but does not have text detection capabilities, it cannot fully replace the OCR engine.

\section{Conclusion}
In this study, we have proposed a novel method to create an individualized preprocessor to improve the accuracy of existing OCR solutions. Two different approaches to the proposed method were tested. One of them, the NN approximation-based training pipeline, was able to improve the performance of two OCR engines on two different datasets, which demonstrates the power and versatility of the proposed preprocessing method based on gradient approximation for OCR tasks. The recognized downside of this approach is the added complexity in the training regimen and the increased number of hyperparameters, and the need for ground truth text with bounding boxes.  Different approximation models can be considered to reduce the need for text bounding boxes in the ground truth as future work. Furthermore, different preprocessor architectures can be investigated in the same setting to see if they yield better results.
\subsubsection*{Acknowledgements}
Authors acknowledge funding support from Intuit Inc. and the University of Alberta, Canada.
%
%
%
\bibliographystyle{splncs04}
\bibliography{mybibliography.bib}

\begin{thebibliography}{10}
\providecommand{\url}[1]{\texttt{#1}}
\providecommand{\urlprefix}{URL }
\providecommand{\doi}[1]{https://doi.org/#1}

\bibitem{Artaud2018}
{Artaud}, C., {Sidère}, N., {Doucet}, A., {Ogier}, J., {Yooz}, V.P.D.: Find
  it! fraud detection contest report. In: ICPR. pp. 13--18 (2018).
  \doi{10.1109/ICPR.2018.8545428}

\bibitem{Bieniecki2007}
{Bieniecki}, W., {Grabowski}, S., {Rozenberg}, W.: Image preprocessing for
  improving ocr accuracy. In: 2007 International Conference on Perspective
  Technologies and Methods in MEMS Design. pp. 75--80 (2007).
  \doi{10.1109/MEMSTECH.2007.4283429}

\bibitem{Buda2019}
Buda, M., Saha, A., Mazurowski, M.A.: Association of genomic subtypes of
  lower-grade gliomas with shape features automatically extracted by a deep
  learning algorithm. Computers in Biology and Medicine  \textbf{109},
  218--225 (6 2019). \doi{10.1016/j.compbiomed.2019.05.002}

\bibitem{Chen2008}
Chen, Q., sen Sun, Q., Heng, P.A., shen Xia, D.: A double-threshold image
  binarization method based on edge detector. Pattern Recognition  \textbf{41},
   1254--1267 (4 2008). \doi{10.1016/j.patcog.2007.09.007}

\bibitem{Chen2019}
Chen, Y., Shao, Y.: Scene text recognition based on deep learning: A brief
  survey. In: ICCSN (2019)

\bibitem{Garain2008}
{Garain}, U., {Jain}, A., {Maity}, A., {Chanda}, B.: Machine reading of
  camera-held low quality text images: An ica-based image enhancement approach
  for improving ocr accuracy. In: ICPR. pp.~1--4 (2008).
  \doi{10.1109/ICPR.2008.4761840}

\bibitem{Graves2006}
Graves, A., Fernández, S., Gomez, F., Schmidhuber, J.: Connectionist temporal
  classification: Labelling unsegmented sequence data with recurrent neural
  networks. In: ACM International Conference Proceeding Series. vol.~148, pp.
  369--376. ACM Press (2006). \doi{10.1145/1143844.1143891}

\bibitem{Harraj2015}
Harraj, A.E., Raissouni, N.: Ocr accuracy improvement on document images
  through a novel pre-processing approach. Signal and Image Processing : An
  International Journal  \textbf{6},  01--18 (9 2015).
  \doi{10.5121/sipij.2015.6401}

\bibitem{He2019}
He, S., Schomaker, L.: Deepotsu: Document enhancement and binarization using
  iterative deep learning. Pattern Recognition  \textbf{91},  379--390 (7
  2019). \doi{10.1016/j.patcog.2019.01.025}

\bibitem{Hicks1979}
Hicks, J.R., J.~C.~Eby, J.: Signal processing techniques in commercially
  available high-speed optical character reading equipment. In: Tao, T.F. (ed.)
  Real-Time Signal Processing II. vol.~0180, pp. 205--216. SPIE (9 1979).
  \doi{10.1117/12.957332}

\bibitem{Hornik1990UniversalAO}
Hornik, K., Stinchcombe, M.B., White, H.: Universal approximation of an unknown
  mapping and its derivatives using multilayer feedforward networks. Neural
  Networks  \textbf{3},  551--560 (1990)

\bibitem{Huang2019}
{Huang}, Z., {Chen}, K., {He}, J., {Bai}, X., {Karatzas}, D., {Lu}, S.,
  {Jawahar}, C.V.: Icdar2019 competition on scanned receipt ocr and information
  extraction. In: ICDAR. pp. 1516--1520 (2019). \doi{10.1109/ICDAR.2019.00244}

\bibitem{Jacovi2019}
Jacovi, A., Hadash, G., Kermany, E., Carmeli, B., Lavi, O., Kour, G., Berant,
  J.: Neural network gradient-based learning of black-box function interfaces.
  In: ICLR. arXiv (1 2019)

\bibitem{Jaderberg14c}
Jaderberg, M., Simonyan, K., Vedaldi, A., Zisserman, A.: Synthetic data and
  artificial neural networks for natural scene text recognition. In: Workshop
  on Deep Learning, NIPS (2014)

\bibitem{lam1995}
Lam, L., Suen, C.Y.: An evaluation of parallel thinning algorithms for
  character recognition. IEEE Transactions on Pattern Analysis and Machine
  Intelligence  \textbf{17},  914--919 (1995). \doi{10.1109/34.406659}

\bibitem{lat2018}
{Lat}, A., {Jawahar}, C.V.: Enhancing ocr accuracy with super resolution. In:
  ICPR. pp. 3162--3167 (2018). \doi{10.1109/ICPR.2018.8545609}

\bibitem{levenshtein1966binary}
Levenshtein, V.I.: Binary codes capable of correcting deletions, insertions,
  and reversals. Soviet physics. Doklady  \textbf{10},  707--710 (1965)

\bibitem{Lillicrap2016}
Lillicrap, T.P., Hunt, J.J., Pritzel, A., Heess, N., Erez, T., Tassa, Y.,
  Silver, D., Wierstra, D.: Continuous control with deep reinforcement
  learning. arXiv preprint arXiv:1509.02971  (2015)

\bibitem{Liu1993}
{Liu}, Y., {Feinrich}, R., {Srihari}, S.N.: An object attribute thresholding
  algorithm for document image binarization. In: ICDAR. pp. 278--281 (1993).
  \doi{10.1109/ICDAR.1993.395732}

\bibitem{Mohamed2019}
Mohamed, S., Rosca, M., Figurnov, M., Mnih, A.: Monte carlo gradient estimation
  in machine learning. Journal of Machine Learning Research  \textbf{21},
  1--63 (6 2019), \url{http://arxiv.org/abs/1906.10652}

\bibitem{Chang1995}
{Moon-Soo Chang}, {Sun-Mee Kang}, {Woo-Sik Rho}, {Heok-Gu Kim}, {Duck-Jin Kim}:
  Improved binarization algorithm for document image by histogram and edge
  detection. In: ICDAR. vol.~2, pp. 636--639 vol.2 (1995).
  \doi{10.1109/ICDAR.1995.601976}

\bibitem{Nguyen2020}
Nguyen, N.M., Ray, N.: End-to-end learning of convolutional neural net and
  dynamic programming for left ventricle segmentation. In: Proceedings of
  Machine Learning Research. vol.~121, pp. 555--569. PMLR (9 2020)

\bibitem{Ogorman1994}
Ogorman, L.: Binarization and multithresholding of document images using
  connectivity. CVGIP: Graphical Models and Image Processing  \textbf{56},
  494--506 (11 1994). \doi{10.1006/cgip.1994.1044}

\bibitem{otsu1979}
Otsu, N.: A threshold selection method from gray-level histograms. IEEE
  transactions on systems, man, and cybernetics  \textbf{9},  62--66 (1979)

\bibitem{Park2019CORDAC}
Park, S., Shin, S., Lee, B., Lee, J., Surh, J., Seo, M., Lee, H.: Cord: A
  consolidated receipt dataset for post-{}ocr parsing. In: Workshop on Document
  Intelligence at NeurIPS 2019 (2019),
  \url{https://openreview.net/forum?id=SJl3z659UH}

\bibitem{Peng2020}
Peng, X., Wang, C.: Building super-resolution image generator for ocr accuracy
  improvement. In: Bai, X., Karatzas, D., Lopresti, D. (eds.) Document Analysis
  Systems. pp. 145--160. Springer International Publishing, Cham (2020)

\bibitem{Reul2018}
{Reul}, C., {Springmann}, U., {Wick}, C., {Puppe}, F.: Improving ocr accuracy
  on early printed books by utilizing cross fold training and voting. In: 2018
  13th IAPR International Workshop on Document Analysis Systems. pp. 423--428
  (2018). \doi{10.1109/DAS.2018.30}

\bibitem{Ronneberger2015}
Ronneberger, O., Fischer, P., Brox, T.: U-net: Convolutional networks for
  biomedical image segmentation. In: Navab, N., Hornegger, J., Wells, W.M.,
  Frangi, A.F. (eds.) MICCAI. pp. 234--241. Springer International Publishing
  (2015)

\bibitem{saha2017}
Saha, P.K., Borgefors, G., {Sanniti di Baja}, G.: Chapter 1 - skeletonization
  and its applications – a review. In: Saha, P.K., Borgefors, G., {Sanniti di
  Baja}, G. (eds.) Skeletonization, pp. 3 -- 42. Academic Press (2017).
  \doi{10.1016/B978-0-08-101291-8.00002-X}

\bibitem{salimans2017}
Salimans, T., Ho, J., Chen, X., Sidor, S., Sutskever, I.: Evolution strategies
  as a scalable alternative to reinforcement learning (2017)

\bibitem{Sauvola2000}
Sauvola, J., Pietikäinen, M.: Adaptive document image binarization. Pattern
  Recognition  \textbf{33},  225--236 (2 2000).
  \doi{10.1016/S0031-3203(99)00055-2}

\bibitem{Shi2017}
Shi, B., Bai, X., Yao, C.: An end-to-end trainable neural network for
  image-based sequence recognition and its application to scene text
  recognition. IEEE Transactions on Pattern Analysis and Machine Intelligence
  \textbf{39},  2298--2304 (11 2017). \doi{10.1109/TPAMI.2016.2646371}

\bibitem{Sporici2020}
Sporici, D., Cuşnir, E., Boiangiu, C.A.: Improving the accuracy of tesseract
  4.0 ocr engine using convolution-based preprocessing. Symmetry  \textbf{12},
  ~715 (5 2020). \doi{10.3390/SYM12050715}

\bibitem{Thompson2016}
{Thompson}, P., {McNaught}, J., {Ananiadou}, S.: Customised ocr correction for
  historical medical text. In: 2015 Digital Heritage. vol.~1, pp. 35--42
  (2015). \doi{10.1109/DigitalHeritage.2015.7413829}

\bibitem{Vo2018}
Vo, Q.N., Kim, S.H., Yang, H.J., Lee, G.: Binarization of degraded document
  images based on hierarchical deep supervised network. Pattern Recognition
  \textbf{74},  568--586 (2 2018). \doi{10.1016/j.patcog.2017.08.025}

\bibitem{White1983}
White, J.M., Rohrer, G.D.: Image thresholding for optical character recognition
  and other applications requiring character image extraction. IBM Journal of
  Research and Development  \textbf{27},  400--411 (1983).
  \doi{10.1147/rd.274.0400}

\bibitem{Williams1992}
Williams, R.J.: Simple statistical gradient-following algorithms for
  connectionist reinforcement learning. Machine Learning  \textbf{8},  229--256
  (5 1992). \doi{10.1007/bf00992696}

\end{thebibliography}

\end{document}